\pdfoutput=1

\documentclass[11pt]{article}

\usepackage[preprint]{acl}

\usepackage{times}
\usepackage{latexsym}

\usepackage[T1]{fontenc}

\usepackage[utf8]{inputenc}

\usepackage{microtype}

\usepackage{inconsolata}

\usepackage{graphicx}
\usepackage{multirow}
\usepackage{enumitem}
\usepackage{array}
\usepackage{algorithm}
\usepackage[noend]{algpseudocode}
\usepackage{float}
\usepackage{colortbl}
\usepackage{amsmath} 
\usepackage{amssymb}
\usepackage{arydshln}
\usepackage{graphics}
\usepackage{booktabs}
\definecolor{done}{rgb}{0.6,0,0.6}

\definecolor{tt}{rgb}{1.0,0.01, 0.24}

\definecolor{cadmiumgreen}{rgb}{0.0, 0.42, 0.24}
\definecolor{new}{rgb}{0,0,1}

\algnewcommand\algorithmicforeach{\textbf{for each}}
\algdef{S}[FOR]{ForEach}[1]{\algorithmicforeach\ #1\ \algorithmicdo}

\title{Efficient Performance Tracking: Leveraging Large Language Models for Automated Construction of Scientific Leaderboards}

\author{Furkan \c{S}ahinu\c{c}$^{1}$, Thy Thy Tran$^{1}$, Yulia Grishina$^{2}$, \\
\textbf{Yufang Hou$^{1,3}$, Bei Chen$^{2}$, Iryna Gurevych$^{1}$} \\
$^{1}$Ubiquitous Knowledge Processing Lab (UKP Lab) \\
Technical University of Darmstadt and Hessian Center for AI (hessian.AI) \\
$^{2}$Amazon Alexa AI - Berlin, Germany \\
$^{3}$IBM Research Europe - Ireland \\
\url{www.ukp.tu-darmstadt.de}}

\begin{document}
\maketitle
\begin{abstract}
Scientific leaderboards are standardized ranking systems that facilitate evaluating and comparing competitive methods. Typically, a leaderboard is defined by a \emph{task}, \emph{dataset}, and evaluation \emph{metric} (TDM) triple, allowing objective performance assessment and fostering innovation through benchmarking. However, the exponential increase in publications has made it infeasible to construct and maintain these leaderboards manually. Automatic leaderboard construction has emerged as a solution to reduce manual labor. Existing datasets for this task are based on the community-contributed leaderboards without additional curation. Our analysis shows that a large portion of these leaderboards are incomplete, and some of them contain incorrect information. In this work, we present \textsc{SciLead}, a manually-curated \textbf{Sci}entific \textbf{Lead}erboard dataset that overcomes the aforementioned problems. Building on this dataset, we propose three experimental settings that simulate real-world scenarios where TDM triples are fully defined, partially defined, or undefined during leaderboard construction. While previous research has only explored the first setting, the latter two are more representative of real-world applications. To address these diverse settings, we develop a comprehensive LLM-based framework for constructing leaderboards. Our experiments and analysis reveal that various LLMs often correctly identify TDM triples while struggling to extract result values from publications. We make our code\footnote{GitHub: \href{https://github.com/UKPLab/arxiv2024-leaderboard-generation}{UKPLab/leaderboard-generation}} and data\footnote{Data: \href{https://tudatalib.ulb.tu-darmstadt.de/handle/tudatalib/4345}{TUdatalib}} publicly available.
\end{abstract}

\section{Introduction}\label{sec:intro}

The comparison of state-of-the-art scientific methods has become a major challenge for the research community due to the rapidly growing number of scientific publications \cite{Landhuis:2016,Bornmann:2021,Mohammad:2020}. For example, around 100 papers are submitted daily to the arXiv pre-print repository under the Computation and Language category\footnote{\url{https://arxiv.org/}} alone. To facilitate monitoring of research progress and comparison of SOTA model performance, scientific leaderboard platforms have been introduced, such as \emph{NLP-progress}\footnote{\url{https://nlpprogress.com/}} or \emph{paperswithcode}\footnote{\url{https://paperswithcode.com/}} (\emph{PwC}). A scientific leaderboard is typically formalized as a \emph{task}, \emph{dataset}, and evaluation \emph{metric} (TDM) triple, ranking performance (\emph{result}) of competitive methods against the triple. However, the majority of these leaderboards are manually curated and maintained, which is infeasible over time due to the ever-expanding number of publications. 

\begin{figure}
    \centering
    \includegraphics[width=0.7\columnwidth]{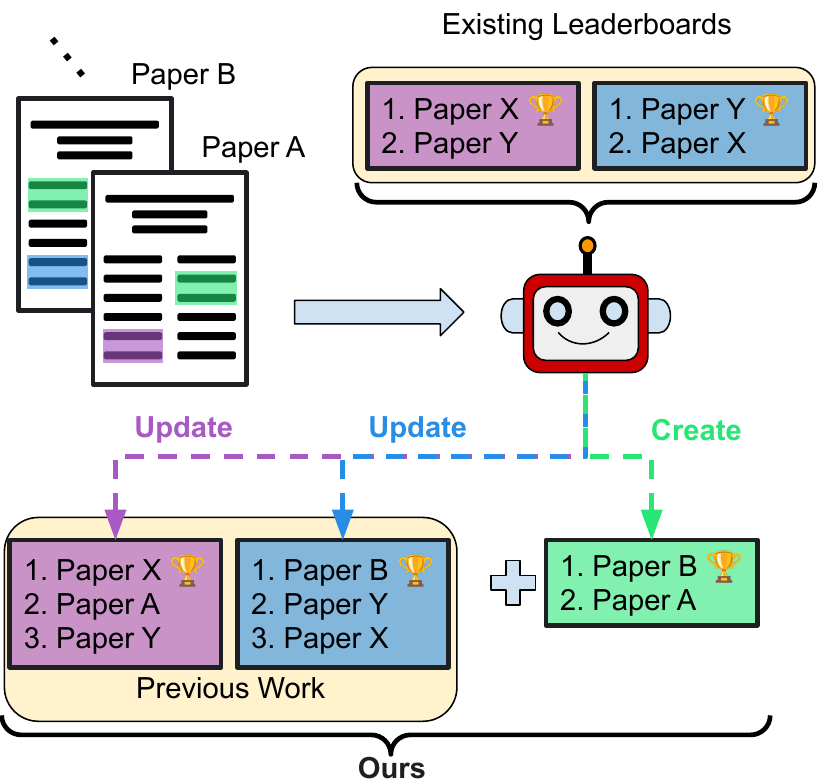}
    \caption{We first extract task, dataset, metric, and result (TDMR) tuples from scientific publications. Then, we update existing leaderboards of the same TDM ({\color{purple}purple} and {\color{blue}blue}). Different from previous work, we also construct a new leaderboard on demand ({\color{cadmiumgreen}green}).} 
    \label{fig:teaser}
\end{figure}

Automatic leaderboard construction has been proposed to offer a solution \cite{Hou:2019}, which aims to extract and compare performance from scientific publications against benchmarks. Previous studies have released datasets based on community-contributed platforms with minimum quality control, such as normalizing the TDM entities across different leaderboards
\cite{Singh:2019,Hou:2019,Hou:2021,Kabongo:2021,Kabongo:2023a,Kabongo:2023b,Yang:2022,Singh:2024}. However, these annotations are limited by individual researchers' varying research interests, resulting in incomplete coverage of papers within a leaderboard and a lack of TDMs from each paper (see Appendix \ref{app:pwc} for a detailed example). Such partial annotations hinder the evaluation of automatic leaderboard construction systems.

Furthermore, all previous work on leaderboard construction mentioned above has formulated the task as matching the extracted TDM triples from a paper against a pre-defined TDM triple taxonomy. This assumption, however, falls short in capturing the dynamic nature of real-world scenarios. In reality, the landscape of leaderboards is constantly evolving, with new ones emerging as a natural consequence of ongoing research and innovation.

To this end, we introduce \textsc{SciLead}, a manually-curated \textbf{Sci}entific \textbf{Lead}erboard dataset, including 27 leaderboards derived from 43 NLP papers. To avoid the pitfall of community-contributed leaderboards with incomplete and inaccurate information, we exhaustively annotate all unique TDM triples and the corresponding results from each paper and construct a \textbf{complete leaderboard set} for these papers. As illustrated in Figure \ref{fig:teaser}, we propose a three-stage framework for constructing leaderboards, harnessing the power of Large Language Models (LLMs) through Retrieval-Augmented Generation (RAG) prompting to streamline the process: (i) extracting task, dataset, metric and result (TDMR) tuples from individual papers; (ii) normalizing extracted TDM triples to the existing pre-defined TDM triple taxonomy or creating new leaderboards; and (iii) ranking papers by their corresponding performance to construct leaderboards.

In the second step, we design three experimental settings that mimic diverse real-world scenarios where we need to update existing leaderboards or create new ones from scratch. Specifically, the TDM triples are either fully specified, partially specified, or completely unknown during the construction process. Notably, the task becomes increasingly challenging in the second and third settings, which have been neglected by all prior work, as it demands reasoning across all existing leaderboards and generating novel ones.

Following previous work, we evaluate TDM extraction using exact match metrics. We further propose various metrics to evaluate the constructed leaderboards, including coverage of correctly assigned publications to the leaderboards and their result values. We also compare the rankings of the gold and constructed leaderboards.

In summary, this work presents \textsc{SciLead}, a manually curated dataset of scientific leaderboards with comprehensive annotations. We propose an LLM-based framework for constructing scientific leaderboards in realistic scenarios. Our experiments demonstrate the competitiveness of our approach in a cold start setting, where we evaluate its performance on four state-of-the-art LLMs. Furthermore, we show that our method can effectively reconstruct scientific leaderboards in real-world scenarios, highlighting its practical applicability.

\section{Related Work}\label{sec:rw}

\paragraph{Scientific Leaderboard Construction.}
Table \ref{tab:related_work} summarizes the differences between previous work and our study. In terms of data source, previous studies use either \textit{NLP-progress} or \textit{paperswithcode}. These sources, however, lack rigorous quality assurance, such as standardizing TDM entities across different leaderboards and ensuring complete coverage of relevant publications. Similar to our work, \citet{Hou:2019} and \citet{Kardas:2020} extract TDM triples along with the results values and apply normalization for leaderboard construction. However, both studies assume a closed domain and match extracted TDM triples to a pre-defined TDM triple taxonomy. On the other hand, some studies only partially extract TDMR tuples and do not apply normalization. For example, \citet{Kabongo:2023b} and \citet{Yang:2022} extract TDM triples without results. Therefore, these works do not deal with leaderboard construction. In addition, \citet{Singh:2024} extract the results values depending on the pre-defined TDM triples. Both \citet{Kabongo:2023b} and \citet{Singh:2024} leverage pre-defined TDM triples in an extraction process similar to \citet{Hou:2019}. Since these approaches require a pre-defined taxonomy of TDM triples, they are incompatible with a realistic task definition. In short, none of the previous work is adaptable to the constantly emerging benchmarks driven by new research and innovation. 
In this work, we address the aforementioned problems. Unlike previous work, we (1) manually construct our dataset directly from publications to ensure complete TDMR annotations, (2) apply normalization for leaderboard construction, and (3) propose different experimental settings to simulate real-world scenarios.

\paragraph{Scientific Knowledge Graph Construction.} Part of the scientific leaderboards can be viewed as a special type of scientific knowledge graph that includes three types of entities (Task, Dataset, Metric) and the relations between them, which have been the primary focus of the previous studies on information extraction from scientific literature \cite{Luan:2018,Jain:2020,Hou:2021,Mondal:2021,Pramanick:2023}. Our work in the cold start scenario, in which we do not assume any pre-defined TDM triple is given, constructs such a scientific knowledge graph and links the papers to the nodes in the graph simultaneously.

\begin{table}[!t]
\centering
\small
\resizebox{1\columnwidth}{!}{
\begin{tabular}{lcccc}
\toprule
\multirow{2}{*}{\textbf{Related Work}}
&\multicolumn{1}{c}{\multirow{2}{*}{\textbf{\begin{tabular}[c]{@{}c@{}}Data\\ Source\end{tabular}}}}
& \multirow{2}{*}{\textbf{Norm}}
& \multicolumn{2}{c}{\textbf{Task Form}}
\\
\cmidrule{4-5}
&&& \textbf{Extr} & \textbf{Constr}
\\
\midrule
\citep{Hou:2019} & NProg. & \checkmark & \checkmark & \checkmark \\
\citep{Kardas:2020} & PwC & \checkmark & \checkmark & \checkmark \\
\citep{Yang:2022}  & PwC & - & $\sim$ & -  \\
\citep{Kabongo:2023b} & PwC & - & $\sim$ & - \\
\citep{Singh:2024} & PwC & - & $\sim$ &  \checkmark  \\
\textsc{SciLead} (Ours) & 
\begin{tabular}[c]{@{}c@{}}Pub.\end{tabular} 
& \checkmark & \checkmark & \checkmark \\
\bottomrule
\end{tabular}
}
\caption{Comparison of related work and ours. \textbf{Data Source}: Source of leaderboards:  NProg.: \emph{NLP-progress}, PwC: \emph{paperswithcode}, Pub.: Publications. \textbf{Norm} refers to Normalization. \textbf{Task Form}: Task formulation, including Extr.: TDMR extraction and Constr.: Leaderboard Construction. $\sim$ indicates partial fulfillment.}
\label{tab:related_work}
\end{table}

\section{Our \textsc{SciLead} Dataset}\label{sec:data}

\begin{figure*}[!th]
    \centering
    \includegraphics[width=0.82\textwidth]{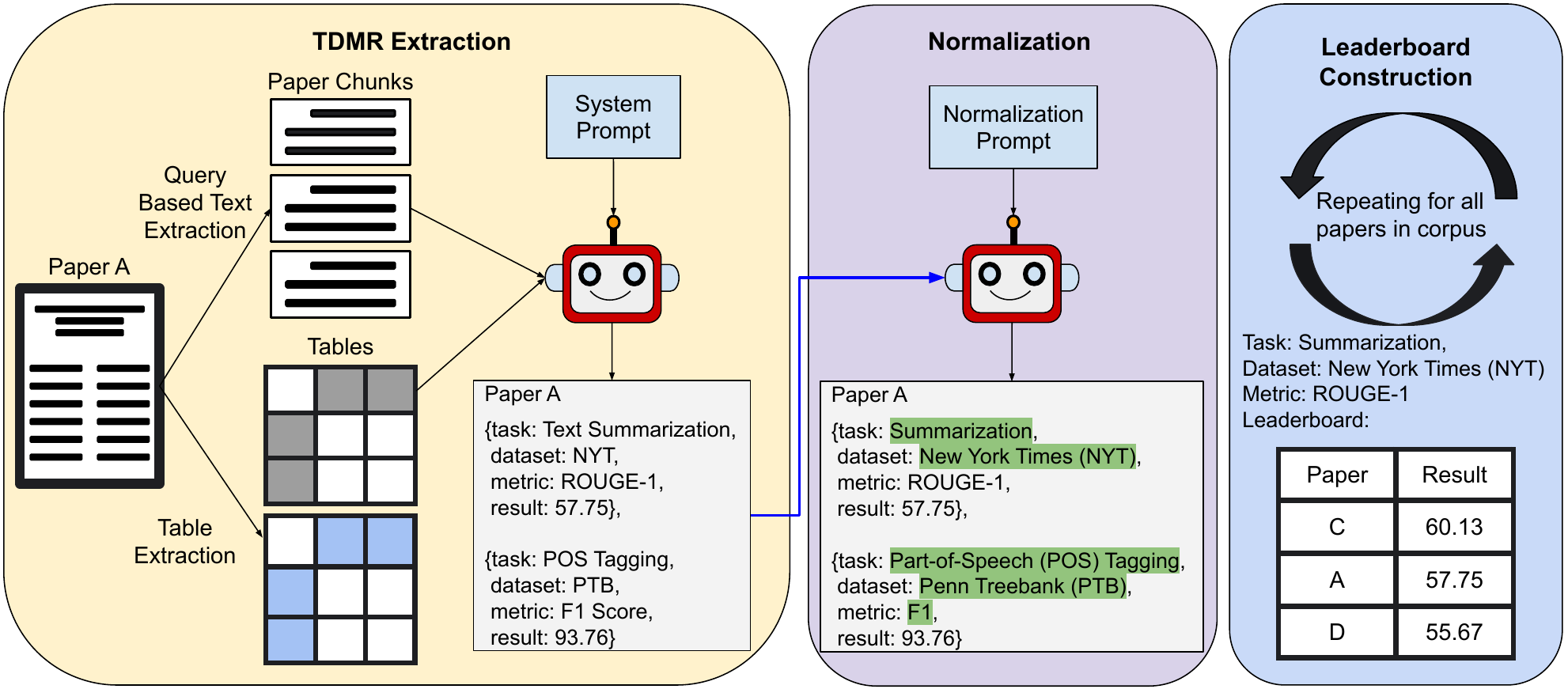}
    \caption{Our framework in three steps: (1) TDMR Extraction, (2) Normalization, (3) Leaderboard Construction}
    \label{fig:single_paper_framework}
\end{figure*}

To facilitate a thorough evaluation of leaderboard construction, 
we require a dataset that satisfies the following criteria. (1) \textbf{TDM coverage}: the dataset should annotate a complete set of all task, dataset, metric (TDM) triples from individual publications, thereby mitigating the problems of incomplete and inaccurate information inherent in previous datasets. (2) \textbf{TDM disambiguation}: mentions of the same TDM entities should be normalized to allow comparison of methods or research in the same leaderboards. (3) \textbf{State-of-the-Art}: for each leaderboard, the associated publications should be ranked based on the best-reported results obtained by their proposed methods (excluding baselines, ablation experiments, or reproduction of other methods), which would restrict each publication to have a single entry per leaderboard. This enables evaluation using ranking similarity metrics. (4) \textbf{Impact}: the dataset should strike a balance between having too many or too few leaderboards, where the former may result in very few papers per each leaderboard, and the latter only constructs popular ones. 

To fulfill these requirements, we created \textsc{SciLead}, a new manually-curated \textbf{Sci}entific \textbf{Lead}erboard dataset. We first selected leaderboards that have a large number of publications from \textit{NLP-progress.} We downloaded PDFs of the relevant publications from arXiv or corresponding venues. We then manually extracted a complete set of TDMR tuples from these publications (\emph{Coverage}), in contrast to previous datasets. Next, we manually normalize mentions of TDM triples which refer to the same entities, building a complete TDM taxonomy for this dataset (\emph{Disambiguation}). For an individual unique TDM triple from a paper, the best-reported performance was kept for the next step (\emph{State-of-the-Art}). Lastly, we aggregate and rank the TDMR tuples from different papers under the corresponding leaderboards defined by TDM triples. After constructing all leaderboards, we filter out leaderboards of less than three entries (\emph{Impact}). We present the dataset statistics in Table \ref{tab:data_stats} and a few data instances in Appendix \ref{app:dataset}.

\begin{table}
\centering
\resizebox{0.55\columnwidth}{!}{
\begin{tabular}{cc}
\toprule
\textbf{Dataset item} & \textbf{Count} \\
\midrule
Papers & 43 \\
Unique tasks & 23 \\
Unique datasets & 71 \\
Unique metrics & 26 \\
Unique TDMs & 138 \\
Leaderboards & 27 \\
Avg. papers / leaderboard & 5.19 \\
\bottomrule
\end{tabular}
}
\caption{Statistics of our \textsc{SciLead} dataset}
\label{tab:data_stats}
\end{table}

\section{Framework}\label{sec:method}

Our proposed framework is illustrated in Figure \ref{fig:single_paper_framework}. The main goal is to automatically construct leaderboards given a collection of scientific publications. The framework first receives as input a set of papers in PDF form and extracts a complete set of TDMR tuples from these papers (\S \ref{ssec:tdmrext}). This first stage is realized by prompting an LLM augmented with a dense retriever. The extracted tuples are then passed into a normalization module in the second stage, which decides whether the TDM triples belong to existing leaderboards or comprise new ones (\S \ref{sec:normalization}). In particular, the normalization module maps these tuples to a pre-defined TDM taxonomy or dynamically updates the taxonomy to integrate new TDM entities. Finally, the corresponding best performances of these normalized TDM triples is being used to rank the competitive methods for leaderboard construction (\S \ref{sec:construction}).

\subsection{TDMR Extraction}\label{ssec:tdmrext}

The goal of TDMR extraction is to obtain a complete set of TDMR tuples from a given set of scientific papers. To address this goal, we implement retrieval-augmented generation (RAG), combining an LLM and a knowledge-based retrieval system. The process involves parsing PDFs and building a vector database. Initially, each publication is parsed using an off-the-shelf PDF processing tool to obtain text and tables within the file. The extracted text is then split into smaller chunks of 2,000 characters, approximately equivalent to 500 tokens. This splitting is crucial to process long documents efficiently. Next, the text chunks and tables are transformed into contextualized embeddings using a pre-trained embedding model. These embeddings are stored systematically in a vector database, which serves as a knowledge base for quick and efficient retrieval in RAG.

After creating the vector database, we implement an embedding similarity retriever that selects candidate chunks potentially containing the desired TDMR tuples. We use the following query for retrieval: ``\texttt{Main task, datasets and evaluation metrics}''. Lastly, we instruct LLMs to extract TDMR tuples from the retrieved chunks by concatenating all retrieved chunks and tables and feeding the combined input into the model. We instruct LLMs to extract TDMR tuples solely from the top-performing results achieved by the proposed methods in the paper, excluding baseline results. The prompts for TDMR extraction and implementation details are provided in Appendix \ref{app:prompts} and in \S\ref{sec:exp_settings}.

\subsection{Normalization} \label{sec:normalization}

Since leaderboards are defined by TDM triplets, papers to be ranked together on the same leaderboard must operate on the same TDMs. However, as mentioned in Section \ref{sec:data}, scientific papers often use different terminology to refer to the same entity (e.g., \emph{Named Entity Recognition} and \emph{Name Tagging}). To ensure the comparability of results from different papers, it is essential to normalize TDM triples. One approach to achieve this is to map entity names extracted from TDM triples to pre-defined, standardized entity names, thereby enabling the comparison of results across papers. While related work assumes the existence of a pre-defined TDM taxonomy, this assumption does not align with actual research practices for two reasons: (1) leaderboard information may not always be available, and (2) it is not possible to map newly introduced task, dataset or metric names to pre-defined sets. Therefore, we apply the normalization process in two more realistic settings simulating scenarios where TDM triples of leaderboards are partially defined or not defined at all beforehand. We name those settings as \emph{partially pre-defined TDM triple normalization} and \emph{cold start normalization}. We test our framework from three normalization perspectives to measure its robustness against real-world use cases with varying difficulty levels.

\paragraph{Fully Pre-defined TDM Triples.} In this setting, we normalize the extracted TDM triples to pre-defined task, dataset, and metric entities. For each entity type $t \in \{\emph{Task, Dataset, Metric}\}$, given the extracted entity mention $l_t$, we use LLMs to associate the extracted name with one of the pre-defined entity names in set $S_t$ which is constructed from our dataset. For simplicity, $S_t$ includes all pre-defined entity names in this setting. 
We discard a TDM triple if any entities cannot be mapped to an existing entity list.
Implementation details and utilized prompts are given in Appendix \ref{app:prompts}.

\begin{algorithm}[!th]
\small
\caption{Partially pre-defined TDM / cold start}\label{alg:partially_masking}
\begin{algorithmic}[1]
    \ForEach{$t \in \{\textit{task}, \textit{dataset}, \textit{metric}\}$}
        \State $S'_t \gets$ Pre-defined taxonomy of type $t$ 
        \State $L_t \gets$ LLM extraction outputs of type $t$ 
        \ForEach{$l_t \in L_t$}
            \State $l_t \gets$ \texttt{LLMNorm} ($l_t, S'_t$)
            \If{ $l_t$ not in $S'_t$}
                \State $S'_t \gets S'_t + \{l_t\}$ 
    \EndIf
    \EndFor
    \EndFor
\end{algorithmic}
\end{algorithm}

\paragraph{Partially Pre-defined TDM Triples.} 
Unlike the previous setting, where the TDM taxonomy is fully pre-defined, here we assume that the TDM taxonomy is partially defined. To simulate this setting, we mask a subset of entities in the fully pre-defined TDM triples. The masked subset represents unseen leaderboards, mimicking the continuous updating and evolution in research. As shown in Algorithm \ref{alg:partially_masking}, we first initialize our new entity name set $S'_t$ by removing the masked entity names $M_t$ from the initial set $S_t$. LLMs first decide whether the given extracted entity name $l_t$ matches with any element in $S'_t$. If the model cannot find a match, our algorithm considers it a newly introduced entity name and adds it to $S'_t$. This process repeats until all extracted entity names are processed. In this framework, we want to simulate the real-world scenario where newly introduced tasks and datasets would create new leaderboards while the old ones will be mapped to existing leaderboards. 

For example, assume that one of the masked leaderboard task names is ``\emph{Named Entity Recognition (NER)}''. If the LLM extraction result is \emph{NER} and no other task names represent this task in $S'_t$ at this point, \emph{NER} will be added to $S'_t$, and other LLM extracted task names related to Named Entity Recognition will be normalized to this entry in the following process. Since \emph{NER} is not directly comparable with the pre-defined leaderboard names, we require an additional normalization step at the TDM triple level to enable direct comparison between newly created leaderboards and gold leaderboards during evaluation, as the new leaderboards may not always utilize the pre-defined TDM triple names. Please refer to Appendix \ref{app:prompts} for more details about the second normalization step.

\paragraph{Cold Start.} In the cold start setting, we do not use any pre-defined entity names to simulate the scenario where no leaderboard exists. We mask all pre-defined entity names in $S_t$ and start the process with an empty set $S'_t$ (i.e., $S'_t = S_t - M_t = \emptyset$). As in the \emph{partially pre-defined TDM triples} setting, $S'_t$ is a dynamic set. We gradually update $S'_t$ by adding new items and processing the entity names individually. For instance, LLM encounters a new entity, ``\textit{Summary Generation}'', and adds it to the empty set $S'_t$. When LLM encounters ``\textit{Document Summarization}'' in the subsequent steps, it associates this new entity with ``\textit{Summary Generation}'' and does not update  $S'_t$ and maps ``\textit{Document Summarization}'' to ``\textit{Summary Generation}''. Similar to the previous setting, after extracting all TDMR tuples from individual papers and normalizing each element with the newly defined TDM entities, we carry out the second normalization step at the TDM triple level.

\subsection{Leaderboard Construction} \label{sec:construction}

In scientific papers, a diverse array of metrics are employed for evaluation, ranging from commonly used measures (e.g., accuracy) to task-specific metrics (e.g., perplexity). While many of these metrics are optimally scaled in [0, 1], others, like root mean squared error, may have arbitrary scales. However, the reported values of these metrics can vary across papers, making it challenging to develop a universal normalization method that can scale all result values to a common range for comparison.

To mitigate this complexity, we have developed a post-processing procedure specifically for percentage-like metrics, which are scaled between 0 and 1, such as F1 score and accuracy, in our current dataset. This procedure begins by identifying whether a predicted metric is percentage-like using a pre-defined list of such metrics. Then, we perform post-processing on the corresponding extracted values. The post-processing involves two steps: first, we clean the extracted results by removing special characters (e.g., \% or $\pm$), as well as mean and standard deviation values if presented. Next, we standardize the reporting style by converting all values to percentages, thereby eliminating inconsistencies that may arise from presenting values as floating points in [0, 1] or as percentages. For metrics with different ranges (e.g., perplexity), we retain the extracted values in their original form.

After this process, we gather the same normalized TDM triples and their corresponding results. We automatically sort result values to determine the rankings of the papers in the same leaderboard. Note that a single paper can belong to multiple leaderboards when working on several task and dataset configurations. As mentioned in Section~\ref{sec:data}, we focus on popular and essential TDMs to keep a practical set of leaderboards. As a result, we only keep leaderboards with at least three TDMR tuples.

\section{Experiments}\label{sec:exps}

\subsection{Experimental Settings} \label{sec:exp_settings}

We used the LangChain\footnote{\url{https://github.com/langchain-ai/langchain}} framework to prompt LLMs for TDMR extraction and normalization. 
We implemented various open and closed source large language models, namely Llama-2 \cite{Touvron:2023}, Llama-3 \cite{MetaAI:2024}, Mixtral \cite{Jiang:2024}, and GPT-4 Turbo \cite{OpenAI:2023}. To ensure a fair evaluation, identical prompts were employed across all models. All prompts are presented in Appendix \ref{app:prompts}. To ensure the robustness of our experimental results in the cold start setting, we run the inference with three different random paper orders and report the average evaluation scores across these runs, thereby mitigating the impact of any particular ordering on the results. 
We used AxCell \cite{Kardas:2020} as a baseline since it is the closest approach to ours.
To better understand our normalization strategy, we further employed a baseline for the normalization step, which normalizes the LLM extracted entities to the pre-defined set by calculating cosine similarity (\textbf{CS}). Note that this baseline is only used in the \emph{fully pre-defined TDM triples, as it requires pre-defined entity names to calculate similarity and is not capable of recognizing newly-emerged entities.} The model and experimental details can be viewed in Appendix \ref{app:exps}.

\subsection{Evaluation Settings}\label{sec:evaluation}

\paragraph{Exact Tuple Match (ETM).} We first investigate whether the models can extract correct tuples of task, dataset, metric, and result  (TDMR) values from publications. We compute recall and precision scores of tuple extraction. The recall metric measures the ratio of gold tuples that are correctly extracted, whereas precision measures how many of the extracted tuples are actually correct. 

\paragraph{Individual Item Match (IIM).} Although the extracted tuples may contain partially correct information, the ETM metric does not recognize or reward such outputs. For example, a TDM triple may be accurate, but \textit{Result} is incorrect. Therefore, we assess the extraction of individual components of TDMR tuples to gain a more nuanced understanding of the model's performance. We employ precision and recall to measure to what extent a model can extract individual items. This allows us to identify local errors in the framework, suggesting potential directions for future work. 

\paragraph{Leaderboard Evaluation.} We employ different metrics for leaderboard evaluation. We first measure \textbf{recall} of leaderboard construction, i.e., the percentage of gold leaderboards that a model correctly identifies. We then evaluate the paper and result coverage of the generated leaderboards. For each gold leaderboard, we compute the ratio of correctly assigned papers and results over the gold ones. Then, we calculate the macro-average of these results as \textbf{paper coverage (PC)} and \textbf{result coverage (RC)}. Lastly, we compare the paper rankings in the gold and model-generated leaderboards using \textbf{Average Overlap (AO)}\footnote{\url{https://github.com/changyaochen/rbo}}~\cite{Webber:2010}. AO can measure the similarity of two ranked leaderboards even when one is incomplete or over-inclusive, such as missing some papers or including wrong papers. This cannot be done if we simply compare the extracted result values. AO is calculated by comparing the overlap between the top-$d$ gold and predicted papers of each leaderboard, denoted as $G_d$ and $M_d$ respectively:

\begin{equation}
A_d = \frac{|G_d \cap M_d|}{d} ;
AO = \frac{1}{k} \sum_{d=1}^k A_d
\end{equation}

\noindent where $A_d$ is agreement between two ranked lists, with $k$ being the length of the shorter leaderboard.

\section{Results and Analysis}\label{sec:results}

In this section, we first inspect the main TDMR extraction and leaderboard construction results obtained under diverse experimental settings on \textsc{SciLead}.
In addition, we augment our analysis by conducting human evaluation and error analysis on a new dataset containing recent NLP papers and papers from the medical domain, showcasing the practical utility of our proposed system in real-world scenarios where TDM triples are not pre-defined.

\begin{table}
\centering
\resizebox{0.65\columnwidth}{!}{
\renewcommand{\arraystretch}{1.05}
\begin{tabular}{ccccc} 
\cline{2-5}
& \multirow{2}{*}{\textbf{Model}} & \multicolumn{3}{c}{\textbf{ETM}} \\
\cline{3-5}
& & \textbf{R} & \textbf{P} & \textbf{F1} \\
\hline
\parbox[t]{2mm}{\multirow{9}{*}{\rotatebox[origin=c]{90}{\textbf{Fully}}}} & AxCell & 13.67 & 32.59 & 19.26 \\
\cdashline{2-5}
& Llama-2 + CS & 21.59 & 10.06 & 13.73 \\
& Llama-2 & 15.25 & 9.63 & 11.81 \\
\cdashline{2-5}
& Mixtral + CS & 24.61 & 26.54 & 25.54 \\
& Mixtral & 21.73 & 24.66 & 23.10 \\
\cdashline{2-5}
& Llama-3 + CS & 29.54 & 23.22 & 26.00 \\
& Llama-3 & 35.60 & 27.11 & 30.78 \\
\cdashline{2-5}
& GPT-4 + CS & 48.71 & 49.82 & 49.26 \\
& GPT-4 & \underline{\textbf{54.53}} & \underline{\textbf{56.02}} & \underline{\textbf{55.27}} \\
\hline
\parbox[t]{2mm}{\multirow{4}{*}{\rotatebox[origin=c]{90}{\textbf{Partially}}}} & Llama-2 & 9.89 & 4.17 & 5.87 \\
& Mixtral & 12.27 & 14.65 & 13.35 \\
& Llama-3 & 18.75 & 15.70 & 17.09 \\
& GPT-4 & \underline{39.56} & \underline{40.60} & \underline{40.07} \\
\hline
\end{tabular}
}
\caption{Exact tuple match (\textbf{ETM}) evaluation scores for different normalization settings (\%). \textbf{R:} Recall, \textbf{P:} Precision, \textbf{F1:} F1 score. LLM + CS indicates the cosine similarity baseline for normalizing individual entities. The best results for each normalization setting are underlined. The overall highest results are bolded.}
\label{tab:etm_scores}
\end{table}

\begin{table*}
\centering
\resizebox{0.8\textwidth}{!}{
\renewcommand{\arraystretch}{1.05}
\begin{tabular}{ccccc!{\vrule width 2pt}ccc!{\vrule width 2pt}ccc!{\vrule width 2pt}ccc} 
\cline{2-14}
& \multirow{2}{*}{\textbf{Model}} & \multicolumn{3}{c!{\vrule width 2pt}}{\textbf{IIM-Task}} & \multicolumn{3}{c!{\vrule width 2pt}}{\textbf{IIM-Dataset}} & \multicolumn{3}{c!{\vrule width 2pt}}{\textbf{IIM-Metric}} & \multicolumn{3}{c}{\textbf{IIM-Result}} \\

\cline{3-14}

& & \textbf{R} & \textbf{P} & \textbf{F1} & \textbf{R} & \textbf{P} & \textbf{F1} & \textbf{R} & \textbf{P} & \textbf{F1} & \textbf{R} & \textbf{P} & \textbf{F1} \\
\hline
\parbox[t]{2mm}{\multirow{9}{*}{\rotatebox[origin=c]{90}{\textbf{Fully}}}} & AxCell & 58.52 & 68.98 & 63.32 & 33.87 & 63.66 & 44.22 & 51.36 & 69.35 & 59.01 & 18.41 & 45.32 & 26.18 \\
\cdashline{2-14}
& Llama-2 + CS & 67.20 & 59.83 & 63.30 & 58.81 & 68.93 & 63.47 & 61.41 & 67.36 & 64.25 & \multirow{2}{*}{31.61} & \multirow{2}{*}{23.75} & \multirow{2}{*}{27.12} \\
& Llama-2 & 60.74 & 55.45 & 57.97 & 55.03 & 62.60 &  	58.57 & 65.49 & 71.51 & 68.37 &  &  & \\
\cdashline{2-14}
& Mixtral + CS & 91.99 & 86.27 & 89.04 & 73.20 & 85.03 & 78.67 & 71.78 & 76.56 & 74.09 & \multirow{2}{*}{41.75} & \multirow{2}{*}{44.62} & \multirow{2}{*}{43.13} \\
& Mixtral & 89.74 & 86.85 & 88.27 & 71.26 & 81.68 & 76.12 & 67.20 & 76.72 & 71.65 &  &  &  \\
\cdashline{2-14}
& Llama-3 + CS & 90.85 & 85.69 & 88.19 & 78.62 & 82.43 & 80.48 & 81.41 & 87.02 & 84.12 & \multirow{2}{*}{49.56} & \multirow{2}{*}{39.50} & \multirow{2}{*}{43.96} \\
& Llama-3 & \underline{\textbf{92.17}} & 87.33 & 89.68 & \underline{\textbf{87.75}} & 92.09 & \underline{\textbf{89.87}} & \underline{\textbf{89.48}} & \underline{\textbf{94.90}} & \underline{\textbf{92.11}} &  &  &  \\
\cdashline{2-14}
& GPT-4 + CS & 90.77 & \underline{\textbf{90.70}} & 90.73 & 79.93 & 86.36 & 83.02 & 81.49 & 86.36 & 83.85 & \multirow{2}{*}{\underline{\textbf{68.22}}} & \multirow{2}{*}{\underline{\textbf{70.34}}} & \multirow{2}{*}{\underline{\textbf{69.26}}} \\

& GPT-4 & 91.10 & 90.62 & \underline{\textbf{90.86}} & 86.05 & \underline{\textbf{92.64}} & 89.22 & 86.46 & 88.18 & 87.31 &  &  &  \\
\hline
\parbox[t]{2mm}{\multirow{4}{*}{\rotatebox[origin=c]{90}{\textbf{Partially}}}} & Llama-2 & 42.98 & 39.70 & 41.27 & 33.14 & 41.05 & 36.67 & 59.34 & 61.24 & 60.28 & 31.61 & 23.75 & 27.12 \\
& Mixtral & 60.72 & 50.23 & 54.98 & 44.45 & 49.67 & 46.92 & 71.19 & 78.72 & 74.77 & 41.75 & 44.62 & 43.13 \\
& Llama-3 & \underline{80.39} & \underline{65.72} & \underline{72.32} & 62.86 & 66.81 & 64.77 & 88.90 & \underline{\textbf{94.90}} & \underline{91.80} & 49.56 & 39.50 & 43.96 \\
& GPT-4 & 78.30 & 63.82 & 70.32 & \underline{79.52} & \underline{83.29} & \underline{81.36} & \underline{89.27} & 92.21 & 90.72 & \underline{\textbf{68.22}} & \underline{\textbf{70.34}} & \underline{\textbf{69.26}} \\
\hline
\end{tabular}
}
\caption{Individual item match (\textbf{IIM}) scores (\%). \textbf{R:} Recall, \textbf{P:} Precision, \textbf{F1:} F1 score. The best results for each setting are underlined. Overall highest results are given in bold. Since normalization is not applied to \emph{Results}, its scores are the same across both settings. The evaluation of the cold start setting is not applicable, as no pre-defined labels are used and only gold leaderboards with a minimum of three entries are used for normalization.}
\label{tab:iim_scores}
\end{table*}

\subsection{Main Results on \textsc{SciLead}}

\paragraph{ETM Evaluation.} Table \ref{tab:etm_scores} shows the ETM scores of different models under two settings: \emph{fully pre-defined TDM triples} and \emph{partially pre-defined TDM triples}.
Note that we do not provide results for the cold start setting, as it does not use pre-defined gold TDM labels and we only normalize generated TDM triples to the gold leaderboards with at least three entries in the second normalization step. We observe that all LLMs, except Llama-2, outperform the AxCell baseline regarding the F1 score in the \emph{fully pre-defined TDM triples} setting. The low precision score of Llama-2 is due to the generation of an excessive number of TDMR tuples. In contrast, AxCell has high precision and low recall compared with Llama-2 due to the small number of extracted tuples. We observe that normalization using LLama-2 and Mixtral falls short of the +CS baseline, whereas Llama-3 and GPT-4 Turbo demonstrate superior performance, outperforming the baseline. As expected, the performance of all models deteriorates in the more realistic, \emph{partially pre-defined TDM triples} setting, which mirrors real-world application scenarios. Nevertheless, GPT-4 Turbo outperforms its counterparts in both settings.

\paragraph{IIM Evaluation.} In Table \ref{tab:iim_scores}, we demonstrate the extraction and normalization performance of different models on individual elements of TDMR tuples under settings with both \emph{fully} and \emph{partially pre-defined TDM triples}. We observe that individual extraction scores are much higher than the exact tuple match scores compared to Table \ref{tab:etm_scores}. It can be concluded that the models are, in fact, effective in the extraction of individual items, but they are struggling to combine those items to create accurate TDMR tuples. Furthermore, all models consistently fall short of extracting correct \emph{Results} compared to other items. This is another significant factor for low ETM scores reported in Table \ref{tab:etm_scores} (further analysis in Appendix \ref{app:etm_scores_wo_results}). Another interesting observation is that Llama-3 is very competitive with GPT-4 Turbo in extracting \emph{Task, Dataset, Metric} items, but its performance considerably drops for extracting the \emph{Result} values.

\begin{table}[h]
\centering
\renewcommand{\arraystretch}{1.05}
\resizebox{0.75\columnwidth}{!}{
\begin{tabular}{cccccc}
\cline{2-6}
& \textbf{Model} & \textbf{LR} & \textbf{PC} & \textbf{RC} & \textbf{AO} \\
\hline
\parbox[t]{2mm}{\multirow{9}{*}{\rotatebox[origin=c]{90}{\textbf{Fully}}}} & AxCell & 40.00 & 25.78 & 15.45 & 11.93 \\
\cdashline{2-6}
& Llama-2 + CS & 70.37 & 35.23 & 8.88 & 4.60 \\
& Llama-2 & 70.37 & 34.96 & 8.26 & 4.96 \\
\cdashline{2-6}
& Mixtral + CS & 96.30 & 53.60 & 20.58 & 13.83 \\
& Mixtral & 88.89 & 46.85 & 16.32 & 11.96 \\
\cdashline{2-6}
& Llama-3 + CS & 81.48 & 60.15 & 22.65 & 21.45 \\
& Llama-3 & 96.30 & \underline{\textbf{79.18}} & 29.30 & 25.49 \\
\cdashline{2-6}
& GPT-4 + CS & 92.59 & 63.70 & 45.59 & 38.66 \\
& GPT-4 & \underline{\textbf{100.00}} & 70.37 & \underline{\textbf{51.79}} & \underline{\textbf{53.87}} \\
\hline
\parbox[t]{2mm}{\multirow{4}{*}{\rotatebox[origin=c]{90}{\textbf{Partially}}}} & Llama-2 & 74.07 & 30.75 & 4.02 & 1.18 \\
& Mixtral & 77.78 & 28.70 & 12.01 & 12.47 \\
& Llama-3 & \underline{92.59} & \underline{61.90} & 19.01 & 19.52 \\
& GPT-4 & 88.89 & 55.92 & \underline{40.06} & \underline{43.71} \\
\hline
\parbox[t]{2mm}{\multirow{4}{*}{\rotatebox[origin=c]{90}{\textbf{Cold Start}}}} & Llama-2 & 16.05 & 5.66 & 0.49 & 0.05 \\
& Mixtral & 49.38 & 20.49 & 8.10 & 3.03 \\
& Llama-3 & 79.01 & \underline{58.78} & 17.18 & 17.63 \\
& GPT-4 & \underline{81.48} & 58.74 & \underline{46.13} & \underline{48.15} \\
\hline
\end{tabular}
}
\caption{Gold leaderboard evaluation (\%). \textbf{LR:} Leaderboard recall, \textbf{PC:} Paper coverage, \textbf{RC:} Result coverage, \textbf{AO:} Average Overlap. The best results for each setting are underlined. Overall best results are given in bold. Standard dev. for cold start are given in Appendix \ref{app:std_dev}.}
\label{tab:leaderboard_results}
\end{table}

\paragraph{Leaderboard Evaluation.} Table \ref{tab:leaderboard_results} shows the leaderboard construction performance under three settings described in Section \ref{sec:normalization}. Although Llama-3 and GPT-4 Turbo are competitive in leaderboard recall ratio, Llama-3 has the best results for the paper coverage across all settings. However, since it struggles with extracting the correct results, as shown in Table \ref{tab:iim_scores}, GPT-4 Turbo takes the lead in result coverage and average scores. Furthermore, we observe that performance drops for most models when the experimental setting becomes more complex and realistic. Interestingly, we found that, unlike other models, GPT-4 performs better in the \emph{cold start }setting compared to the \emph{partially pre-defined TDM triples} setting on paper and result coverage, as well as average overlap. We provide further analysis of this behavior in the next section.

\subsection{Partial Pre-defined Leaderboard Analysis}

Table \ref{tab:masked_leaderbaord_results} shows the comparison results of different models on the  \emph{partially pre-defined TDM triples} and \emph{cold start} settings on the masked leaderboard subset in the \emph{partially pre-defined TDM triples} setting (based on $M_t$ in Algorithm \ref{alg:partially_masking}). Overall, Llama-2 and Mixtral perform better in the \emph{partially pre-defined TDM triples} setting, while Llama-3 and GPT-4 Turbo perform better in a cold start setting. When we look at the different results for these two settings, we notice that in the \emph{partially pre-defined TDM triples} setting, sometimes the model makes errors in the normalization step by mapping a newly extracted TDM triple to an existing pre-defined TDM triple that has the similar surface form but refers to a different leaderboard. For instance, the model maps \emph{CoNLL-2003 - English} to \emph{CoNLL-2003 - German}. Such errors occur less frequently in the \emph{cold start} setting. 

\begin{table}
\centering
\resizebox{0.75\columnwidth}{!}{
\renewcommand{\arraystretch}{1.05}
\begin{tabular}{cccccc}
\toprule
& \textbf{Model} & \textbf{LR} &\textbf{PC} & \textbf{RC} & \textbf{AO} \\
\hline
\parbox[t]{2mm}{\multirow{4}{*}{\rotatebox[origin=c]{90}{\textbf{Partially}}}} & Llama-2 & 60.00 & 17.76 & 6.71 & 2.08 \\
& Mixtral & 60.00 & 20.26 & 13.26 & 20.28 \\
& Llama-3 & 90.00 & 57.38 & 11.83 & 7.01 \\
& GPT-4 & 70.00 & 50.57 & 33.50 & 26.03 \\
\hline
\parbox[t]{2mm}{\multirow{4}{*}{\rotatebox[origin=c]{90}{\textbf{Cold Start}}}} & Llama-2 & 10.00 & 3.10 & 0.67 & 0.00 \\
& Mixtral & 53.33 & 22.67 & 10.97 & 7.91 \\
& Llama-3 & 86.67 & 78.14 & 22.59 & 14.79 \\
& GPT-4 & 83.33 & 60.76 & 43.11 & 37.50 \\
\bottomrule
\end{tabular}
}
\caption{Leaderboard evaluation only on masked leaderboards from partially masking setting (\%). Metrics are the same as Table \ref{tab:leaderboard_results}. Standard dev. for cold start are given in Appendix \ref{app:std_dev}.}
\label{tab:masked_leaderbaord_results}
\end{table}

\subsection{Additional Analysis}
\paragraph{Leaderboard Threshold.} The minimum number of papers required for a leaderboard directly impacts the performance of leaderboard construction, as it influences the total number of leaderboards. In our main experiments, we focused on leaderboards with a minimum of three papers, but we also explored the effects of reducing this threshold to two papers. Notably, this change significantly increased the number of leaderboards from 27 to 62. Table \ref{tab:leaderboard_results_ablation} demonstrates the leaderboard construction results, and all other evaluation results can be seen in Appendix \ref{app:threshold}. In general, the overall performance drops compared to the main results in Table \ref{tab:leaderboard_results}. This is expected due to the increased number of leaderboards. Notably, most of our key findings from the main experiments remain consistent. However, one notable exception is the performance of GPT-4, which excels in the partially pre-defined setting compared to the cold start setting, contrary to our main results. We attribute this discrepancy to the fact that the cold start setting is more severely disadvantaged in the second normalization step, which becomes increasingly challenging as the number of leaderboards grows. In other words, the cold start setting struggles to effectively distinguish between similar TDM pairs in the second normalization step, leading to this observed performance difference.

\paragraph{Zero-shot vs Few-shot.} While zero-shot prompting LLMs achieves impressive performance in leaderboard construction, we additionally explore whether including few-shot examples in the prompt can further enhance the results. To this end, we conducted experiments using the best model (GPT-4 Turbo) in the \emph{fully pre-defined} setting. We selected a paper not included in \textsc{SciLead} but working on similar tasks as an example. We provided text chunks, including TDM information, along with the results tables and the expected results for TDMR extraction. For normalization, we used the same pre-defined entity set. The remaining experimental details were identical to the main experiments. Unlike previous work, where few-shot prompting often yields performance gains, our results in Table \ref{tab:few_shot_scores} show that few-shot prompting leads to a slight performance drop compared to zero-shot prompting for exact tuple match (ETM). In contrast, for individual item match (IIM), the few-shot approach outperforms zero-shot prompting for task and metric extraction but underperforms for dataset and result extraction. Future research can explore better strategies for selecting few-shot examples to incorporate into the prompt \cite{Wu:2023}.

\begin{table}
\centering
\small
\renewcommand{\arraystretch}{1.05}
\resizebox{0.88\columnwidth}{!}{
\begin{tabular}{ccccccc} 
\toprule
\textbf{Evaluation} & \textbf{R} & $\Delta_R$ &\textbf{P} & $\Delta_P$ & \textbf{F1} & $\Delta_{F1}$ \\
\midrule
ETM & 51.03 & -3.50 & 51.01 & -5.01 & 51.02 & -4.25 \\
IIM - Task & 92.66 & 1.56 & 92.11 & 1.49 & 92.39 & 1.53 \\
IIM - Dataset & 85.55 & -0.5 & 89.06 & -3.58 & 87.27 & -1.95 \\
IIM - Metric & 93.39 & 6.93 & 95.74 & 7.56 & 94.55 & 7.24 \\
IIM - Result & 67.78 & -0.44 & 69.21 & -1.13 & 68.48 & -0.78 \\
\bottomrule
\end{tabular}
}
\caption{GPT-4 few-shot ETM and IIM results for \emph{fully pre-defined} setting and delta values for zero-shot. }
\label{tab:few_shot_scores}
\end{table}

\subsection{Results on Wild Datasets}

To get further insights into the performance of our framework in a real-world environment, we applied our framework to recent papers from two different domains (NLP and medical) and conducted manual analysis. We first collected 339 main conference and findings papers from the recent EACL 2024 and sampled 12 papers from the medical domain from \emph{paperswithcode}. We sampled papers from the most popular leaderboards in the medical domain and manually checked whether their TDMR tuples matched the leaderboards. We performed the cold start setting with the best model (GPT-4) and gathered the same TDM triples to construct leaderboards from these papers. We sampled the most crowded 24 leaderboards and manually evaluated the correctness of TDMR tuples and leaderboards. Table \ref{tab:human_eval} demonstrates that the task, dataset, and metric information is extracted and normalized with a high success rate. In contrast, there is significant room for improvement in the result extraction. Nevertheless, our framework can extract and normalize TDM triples, leading to high accuracy in assigning papers to the leaderboards across two domains. This emphasizes the generalizability of our framework for real-world scenarios.

\begin{table}
\centering
\small
\resizebox{0.9\columnwidth}{!}{
\begin{tabular}{cccccc}
\hline
\textbf{Domain} & \textbf{Task} & \textbf{Dataset} & \textbf{Metric} & \textbf{Result} & \textbf{CP} \\
\hline
NLP & 0.76 & 0.88 & 0.88 & 0.44 & 0.59 \\
Med. & 1.00 & 1.00 & 1.00 & 0.56 & 1.00 \\
All & 0.84 & 0.92 & 0.92 & 0.48 & 0.72 \\
\hline
\end{tabular}
}
\caption{Ratio of correct individual TDMR items in manual evaluated leaderboards in both domains. \textbf{CP}: Correctly assigned papers to leaderboards.}
\label{tab:human_eval}
\end{table}

\subsection{Error Analysis}\label{sec:error} 

We performed a thorough error analysis from multiple angles. Given that extracting \textit{Results} is the most challenging step for models, we identified the five papers with the lowest GPT-4 \textit{IIM-Results} scores. Our analysis of 58 erroneous \textit{Results} instances revealed three primary error categories. Firstly, 19 errors (33\%) occurred due to confusion with values in other tables, such as failing to extract the best results or mistakenly swapping dev and test sets. Secondly, 18 errors (31\%) arose from confusion with the appendix material. Finally, the remaining 21 errors (36\%) involved missed extractions, often accompanied by errors in other TDM components.

In addition, we also checked three leaderboards that have the lowest paper coverage scores and are constructed in the \textit{partially pre-defined} setting. We observe that LLMs have difficulties normalizing task names that do not have similar surface forms to gold task names. For instance,  ``\textit{Named Entity Recognition}'' is incorrectly mapped to ``\textit{Syntactic Information Extraction}''. We also observe errors in joint task names like \textit{Intent Detection and Slot Filling}. Models may extract task names separately and not normalize them to the joint form.  

Furthermore, we observe that some errors can be attributed to the absence of explicit task or dataset names in the paper. For example, some papers introduce the tasks as \textit{GLUE Benchmark} instead of referring to the original task names such as \textit{Sentiment Analysis}. This prevents the TDMR tuples from being associated with the correct leaderboards. Lastly, some errors occur due to the failure to retrieve related text and tables from the papers.

\section{Conclusion}\label{sec:conc}

We introduce \textsc{SciLead}, a new dataset addressing the limitations of existing leaderboard construction datasets by providing accurate and complete Task-Dataset-Metric-Result information from scientific papers. Furthermore, we develop an LLM-based framework to facilitate the automatic leaderboard construction process in realistic application scenarios. Our results show that LLMs excel in extracting task, dataset, and metric names and detecting correct leaderboards but struggle with extracting result values. Our work contributes to the automatic leaderboard construction problem, providing valuable insights for large-scale applications.

\section{Limitations}

Due to the time-consuming nature of manual extraction, the \textsc{SciLead} dataset is currently limited to a set of 43 papers. However, we believe that \textsc{SciLead} outweighs other large datasets in terms of content because it accurately captures all the unique TDMR tuples in the papers and provides a normalization between TDMRs from different papers. We leave large-scale experiments for follow-up research.

We acknowledge that our dataset is predominantly comprised of papers from the machine learning domain, which offers a rich source of comparable performance scores on specific topics, making it an attractive focus for leaderboard construction studies. Additionally, given that the majority of articles in the literature are published in English, our dataset accordingly consists of English-language articles. However, we recognize the importance of expanding our framework to encompass papers from diverse scientific domains and languages, which presents a promising direction for future research.

\section*{Ethics Statement}

The primary purpose of automatically generating leaderboards is to provide researchers with a concise and comprehensive overview of the scientific literature landscape, facilitating the comparison of previous and current state-of-the-art approaches to a specific task. Notably, all models and data instances utilized in our work, with the exception of GPT-4 Turbo, are built upon open-source foundations, thereby promoting responsible, reproducible, and transparent scientific research practices. Furthermore, our study does not rely on crowd-sourced human annotators.

\section*{Acknowledgments}

This work was funded by the "Modeling Task-oriented Dialogues Grounded in Scientific Literature" project in partnership
with Amazon Alexa. We gratefully acknowledge the support of Microsoft with a grant for access to OpenAI GPT models via the Azure cloud (Accelerate Foundation Model Academic Research). Yufang Hou is supported by the Visiting Female Professor Programme from the Technical University of Darmstadt.

\bibliography{custom}

\appendix

\newpage

\section{Papers with Code Analysis}\label{app:pwc}

We take \cite{Xin:2018} as an example paper. Its TDMR tuples are given in Table \ref{tab:pwc_analysis}. However, its \emph{paperswithcode} entry\footnote{\url{https://paperswithcode.com/paper/learning-better-internal-structure-of-words}} lacks of results for German, Spanish and Dutch datasets. In addition, the dataset for chunking task is confused with Penn Treebank. 

\begin{table}[h]
\centering
\renewcommand{\arraystretch}{1.1}
\resizebox{1\columnwidth}{!}{
\begin{tabular}{cccc}
\toprule
\textbf{Task} & \textbf{Dataset} & \textbf{Metric} & \textbf{Result} \\
\hline
Named Entity Recognition & CoNLL-2003 - English & F1 & 91.64 \\
Named Entity Recognition& CoNLL-2003 - German &  F1 & 79.43 \\
Named Entity Recognition & CoNLL-2002 - Spanish & F1 & 86.68 \\
Named Entity Recognition & CoNLL-2002- Dutch & F1 & 87.81 \\
Text Chunking & CoNLL-2000 & F1 & 95.29 \\
Part-of-Speech Tagging & Penn Treebank (PTB) & F1 & 97.58 \\
\hline
\end{tabular}
}
\caption{TDMR tuples from \cite{Xin:2018}}
\label{tab:pwc_analysis}
\end{table}

\section{Dataset}\label{app:dataset}

We present some representative data instances from \textsc{SciLead} in Table \ref{tab:tdmr_data_sample} and Table \ref{tab:leaderboard_sample} for TDMR tuples and leaderboards respectively.

\begin{table*}
\centering
\small
\renewcommand{\arraystretch}{1.2}
\begin{tabular}{c|c|c|c|c}
\hline
\textbf{Paper Name} & \textbf{Task} & \textbf{Dataset} & \textbf{Metric} & \textbf{Result} \\
\hline
\rowcolor{lightgray} 1703.06345.pdf & Named Entity Recognition (NER) & CoNLL-2003 - English & F1 & 91.26 \\
\rowcolor{lightgray} 1703.06345.pdf & Named Entity Recognition (NER) & CoNLL-2002 - Spanish & F1 & 85.77 \\
\rowcolor{lightgray} 1703.06345.pdf & Named Entity Recognition (NER) & CoNLL-2002- Dutch & F1 & 85.19 \\
\rowcolor{lightgray} 1703.06345.pdf & Text Chunking & CoNLL-2000 & F1 & 95.41 \\
\rowcolor{lightgray} 1703.06345.pdf & Part-of-Speech (POS) Tagging & Penn Treebank (PTB) & Accuracy & 97.55 \\
\hline
\rowcolor{lime} 1603.01354.pdf & Named Entity Recognition (NER) & CoNLL-2003 - English & Precision & 91.35 \\
\rowcolor{lime} 1603.01354.pdf & Named Entity Recognition (NER) & CoNLL-2003 - English & Recall & 91.06 \\
\rowcolor{lime} 1603.01354.pdf & Named Entity Recognition (NER) & CoNLL-2003 - English & F1 & 91.21 \\
\rowcolor{lime} 1603.01354.pdf & Part-of-Speech (POS) Tagging & Penn Treebank (PTB) & Accuracy & 97.55 \\
\hline
\rowcolor{orange} 1709.04109.pdf & Named Entity Recognition (NER) & CoNLL-2003 - English & F1 & 91.85 \\
\rowcolor{orange} 1709.04109.pdf & Text Chunking & CoNLL-2000 & F1 & 96.13 \\
\rowcolor{orange} 1709.04109.pdf & Part-of-Speech (POS) Tagging & Penn Treebank (PTB) & Accuracy & 97.59 \\
\end{tabular}
\caption{Example TDMR instances from \textsc{SciLead}. TDMR tuples per each paper are color coded.}
\label{tab:tdmr_data_sample}
\end{table*}

\begin{table*}
\small
\centering
\renewcommand{\arraystretch}{1.1}
\begin{tabular}{l|ccl|c}
\cline{1-2} \cline{4-5}
\textbf{Task:} Named Entity Recognition (NER) & \multirow{3}{*}{\textbf{Result}} & & \textbf{Task:} Part-of-Speech (POS) Tagging & \multirow{3}{*}{\textbf{Result}} \\
\textbf{Dataset:} CoNLL-2003 - English & & & \textbf{Dataset:} Penn Treebank (PTB) & \\
\textbf{Metric:} F1 & & & \textbf{Metric:} Accuracy & \\
\cline{1-2} \cline{4-5}
\cellcolor{orange} 1709.04109.pdf & \cellcolor{orange} 91.85 &  & \cellcolor{orange} 1709.04109.pdf & \cellcolor{orange} 97.59 \\
\cellcolor{lightgray} 1703.06345.pdf & \cellcolor{lightgray} 91.26 &  & \cellcolor{lime} 1603.01354.pdf & \cellcolor{lime} 97.55 \\
\cellcolor{lime} 1603.01354.pdf & \cellcolor{lime} 91.21 &  & \cellcolor{lightgray} 1703.06345.pdf & \cellcolor{lightgray} 97.55 \\
\end{tabular}
\caption{Example leaderboards from \textsc{SciLead} based on common TDMR tuples in Table \ref{tab:tdmr_data_sample}. The same color codes are used.}
\label{tab:leaderboard_sample}
\end{table*}

\section{Experimental Details} \label{app:exps}

In the TDMR extraction step, Unstructured\footnote{\url{https://github.com/Unstructured-IO/unstructured}} and Chroma\footnote{\url{https://github.com/chroma-core/chroma}} libraries were used for PDF parsing and vector database implementation. \texttt{multi-qa-mpnet-base-dot-v1}~\cite{Reimers:2019} was used as the embedding model for dense retrieval. 

All experiments with Mixtral, Llama-2, and Llama-3 used the instruction-tuned version and were run in 4-bit quantization on a single NVIDIA A100 GPU with 80GB memory, taking approximately $\sim$13 hours in total. We used the Llama-2 Chat 70B, Mixtral-8x7B-Instruct, Llama-3 Instruct 70B  versions, and the API version  \texttt{1106-preview} for GPT-4 Turbo. For reproducibility, we used greedy decoding for all models. In prompting, the only modifications made across the models were to conform to each model's specific format requirements, such as incorporating model-specific tokens.

For embedding similarity model for normalization similarity baseline, we utilized \texttt{bilingual-embedding-large} \cite{Thakur:2021} because of its high performance in the Massive Text Embedding Benchmark (MTEB) \cite{Muennighoff:2023} for Semantic Textual Similarity (STS).

Unlike our framework, AxCell works on TeX files instead of PDFs. 
Therefore, we manually downloaded 36 TeX files from the 43 papers in our dataset. This poses an input constraint to their method. Performance measurements of AxCell for TDMR and leaderboard evaluation were made according to the gold labels of those 36 papers. To normalize AxCell outputs, we applied the \textit{fully pre-defined TDM triples} setting using GPT-4 Turbo.

Since some metrics are common across many tasks and datasets in leaderboards, only some items in the task and dataset are masked in the partially masking setting to avoid making the partially pre-defined setting too close to the cold start.

\paragraph{Evaluation.} We calculate exact tuple match scores as follows: 

\begin{equation}
{ETM}^{Recall} = \frac{1}{|P|} \sum_{p \in P} \frac{|T_p^G \cap T_p^M|}{|T_p^G|}
\end{equation}

\begin{equation}
{ETM}^{Precision} = \frac{1}{|P|} \sum_{p \in P} \frac{|T_p^G \cap T_p^M|}{|T_p^M|}
\end{equation}

\noindent where $p$ represent a single paper in the set of all papers $P$ in the dataset. $T_p^G$ and $T_p^M$ are gold and model generated tuple set for the paper $p$. 

For paper item type $t$, we calculate individual item match scores as follows:

\begin{equation}
IIM^{Recall}_{t} = \frac{1}{|P|} \sum_{p \in P} \frac{|I_{t,p}^G \cap I_{t,p}^M|}{|I_{t,p}^G|},
\end{equation}

\begin{equation}
IIM^{Precision}_{t} = \frac{1}{|P|} \sum_{p \in P} \frac{|I_{t,p}^G \cap I_{t,p}^M|}{|I_{t,p}^M|},
\end{equation}

\noindent where $I_{t,p}^G$ and $I_{t,p}^M$ represent the sets of gold and model generated unique type $t$ items that belong to the paper $p$.

\section{ETM without \emph{Result}.}\label{app:etm_scores_wo_results}

Since extracting the correct results from the papers is the main bottleneck for exact match, we calculate ETM score by excluding \emph{Result} (i.e., on TDMs, not TDMRs) in Table \ref{tab:etm_scores_wo_results}. We see that the scores of all models show a remarkable increase. In this setting, Llama-3 outperforms GPT-4 Turbo in the \emph{fully pre-defined TDM triples} setting, but GPT-4 Turbo is more robust in the \emph{partially pre-defined TDM triples} setting.

\begin{table}
\centering
\small
\renewcommand{\arraystretch}{1.2}
\resizebox{1\columnwidth}{!}{
\begin{tabular}{ccccc}
\cline{2-5}
& \textbf{Model} & \textbf{R (\% change)} & \textbf{P (\% change)} & \textbf{F1 (\% change)} \\
\hline
\parbox[t]{2mm}{\multirow{9}{*}{\rotatebox[origin=c]{90}{\textbf{Fully}}}} & AxCell & 25.78 (\textcolor{teal}{88.59}) & 55.08 (\textcolor{teal}{69.01}) & 35.12 (\textcolor{teal}{82.35})\\
\cdashline{2-5}
& Llama-2 + CS & 39.54 (\textcolor{teal}{83.14}) & 40.66 (\textcolor{teal}{304.17}) & 40.09 (\textcolor{teal}{191.99}) \\
& Llama-2 & 34.91 (\textcolor{teal}{128.92}) & 35.85 (\textcolor{teal}{272.27}) & 35.37 (\textcolor{teal}{199.49}) \\
\cdashline{2-5}
& Mixtral + CS & 52.05 (\textcolor{teal}{111.50}) & 58.44 (\textcolor{teal}{120.20}) & 55.06 (\textcolor{teal}{115.58})\\
& Mixtral & 48.75 (\textcolor{teal}{124.34}) & 55.66 (\textcolor{teal}{125.71}) & 51.97 (\textcolor{teal}{124.98}) \\
\cdashline{2-5}
& Llama-3 + CS & 58.07 (\textcolor{teal}{96.58}) & 62.02 (\textcolor{teal}{167.10}) & 59.98 (\textcolor{teal}{130.69}) \\
& Llama-3 & \underline{\textbf{72.56}} (\textcolor{teal}{103.82}) & \underline{\textbf{77.13}} (\textcolor{teal}{184.51}) & \underline{\textbf{74.77}} (\textcolor{teal}{142.92}) \\
\cdashline{2-5}
& GPT-4 + CS & 63.82 (\textcolor{teal}{31.02}) & 68.95 (\textcolor{teal}{38.40}) & 66.29 (\textcolor{teal}{34.57})\\
& GPT-4 & 70.40 (\textcolor{teal}{29.10}) & 75.28 (\textcolor{teal}{34.38}) & 72.75 (\textcolor{teal}{31.63}) \\
\hline
\parbox[t]{2mm}{\multirow{4}{*}{\rotatebox[origin=c]{90}{\textbf{Partially}}}} & Llama-2 & 22.99 (\textcolor{teal}{132.45}) & 27.23 (\textcolor{teal}{553.00}) & 24.93 (\textcolor{teal}{324.70}) \\
& Mixtral & 24.48 (\textcolor{teal}{99.51}) & 27.89 (\textcolor{teal}{90.38}) & 26.07 (\textcolor{teal}{95.28}) \\
& Llama-3 & 45.30 (\textcolor{teal}{141.60}) & 50.75 (\textcolor{teal}{223.25}) & 47.87 (\textcolor{teal}{180.11}) \\
& GPT-4 & \underline{51.89} (\textcolor{teal}{31.17}) & \underline{56.08} (\textcolor{teal}{38.13}) & \underline{53.90} (\textcolor{teal}{34.51}) \\
\hline
\end{tabular}
}
\caption{ETM scores without \emph{Result} (\%). Percentage change relative to Table \ref{tab:etm_scores} are given in parenthesis highlighted in \textcolor{teal}{teal}. The best results of each normalization setting are underlined. Overall best results are given in bold.}
\label{tab:etm_scores_wo_results}
\end{table}

{\section{Paper Threshold Results}\label{app:threshold}}

We provide full results for leaderboard threshold of two papers in Tables \ref{tab:etm_scores_ablation}, \ref{tab:iim_scores_ablation}, \ref{tab:leaderboard_results_ablation} and \ref{tab:masked_leaderbaord_results_ablation}.

\begin{table}
\centering
\small
\renewcommand{\arraystretch}{1.1}
\resizebox{1\columnwidth}{!}{
\begin{tabular}{cccc|ccc} 
\hline
\multirow{2}{*}{\textbf{Model}} & \multicolumn{3}{c|}{\textbf{ETM}} & \multicolumn{3}{c}{\textbf{ETM w/o \textit{Result}}} \\
\cline{2-7}
& \textbf{R} & \textbf{P} & \textbf{F1} & \textbf{R} & \textbf{P} & \textbf{F1} \\
\hline
Llama-2 & 7.33 & 4.33 & 5.45 & 26.60 & 33.27 & 29.56 \\
Mixtral & 12.49 & 13.45 & 12.95 & 27.85 & 29.87 & 28.83 \\
Llama-3 & 22.82 & 19.04 & 20.78 & 53.42 & 58.84 & 56.00 \\
GPT-4 & 45.47 & 46.90 & 46.17 & 60.24 & 64.43 & 62.26 \\
\hline
\end{tabular}
}
\caption{\textbf{ETM} scores (\%) with and without \textit{Result} values in partially pre-defined setting on the leaderboard dataset with paper threshold two.}
\label{tab:etm_scores_ablation}
\end{table}

\begin{table}
\centering
\renewcommand{\arraystretch}{1.1}
\resizebox{1\columnwidth}{!}{
\begin{tabular}{cccc!{\vrule width 2pt}ccc!{\vrule width 2pt}ccc!} 
\hline
\multirow{2}{*}{\textbf{Model}} & \multicolumn{3}{c!{\vrule width 2pt}}{\textbf{IIM-Task}} & \multicolumn{3}{c!{\vrule width 2pt}}{\textbf{IIM-Dataset}} & \multicolumn{3}{c!}{\textbf{IIM-Metric}} \\
\cline{2-10}
& \textbf{R} & \textbf{P} & \textbf{F1} & \textbf{R} & \textbf{P} & \textbf{F1} & \textbf{R} & \textbf{P} & \textbf{F1} \\
\hline
Llama-2 & 42.98 & 39.70 & 41.27 & 33.14 & 41.05 & 36.67 & 59.34 & 61.24 & 60.28 \\
Mixtral & 60.72 & 50.23 & 54.98 & 44.45 & 49.67 & 46.92 & 71.19 & 78.72 & 74.77 \\
Llama-3 & \underline{80.39} & \underline{65.72} & \underline{72.32} & 62.86 & 66.81 & 64.77 & 88.90 & \underline{\textbf{94.90}} & \underline{91.80} \\
GPT-4 & 78.30 & 63.82 & 70.32 & \underline{79.52} & \underline{83.29} & \underline{81.36} & \underline{89.27} & 92.21 & 90.72 \\
\hline
\end{tabular}
}
\caption{\textbf{IIM} scores (\%) in partially pre-defined setting on the leaderboard dataset with paper threshold two.}
\label{tab:iim_scores_ablation}
\end{table}

\begin{table}
\centering

\renewcommand{\arraystretch}{1.05}
\resizebox{0.75\columnwidth}{!}{
\begin{tabular}{cccccc}
\toprule
& \textbf{Model} & \textbf{LR} & \textbf{PC} & \textbf{RC} & \textbf{AO} \\
\hline
\parbox[t]{2mm}{\multirow{9}{*}{\rotatebox[origin=c]{90}{\textbf{Fully}}}} & AxCell & 32.26 & 13.31 & 9.37 & 9.50 \\
\cdashline{2-6}
& Llama-2 + CS & 46.77 & 24.21 & 4.67 & 2.00 \\
& Llama-2 & 51.61 & 25.70 & 5.21 & 3.77 \\
\cdashline{2-6}
& Mixtral + CS & 66.13 & 38.66 & 11.38 & 6.83 \\
& Mixtral & 62.90 & 36.53 & 10.33 & 8.44 \\
\cdashline{2-6}
& Llama-3 + CS & 56.45 & 41.52 & 11.47 & 9.75 \\
& Llama-3 & 64.52 & 55.44 & 15.18 & 11.91 \\
\cdashline{2-6}
& GPT-4 + CS & 72.58 & 49.51 & 35.98 & 36.18 \\
& GPT-4 & 85.48 & 57.26 & 43.52 & 51.28 \\
\hline
\parbox[t]{2mm}{\multirow{4}{*}{\rotatebox[origin=c]{90}{\textbf{Partially}}}} & Llama-2 & 40.32 & 23.54 & 2.93 & 2.18 \\
& Mixtral & 53.22 & 25.89 & 8.37 & 6.50 \\
& Llama-3 & 74.19 & 53.95 & 12.51 & 14.14 \\
& GPT-4 & 67.74 & 48.47 & 38.65 & 44.10 \\
\hline
\parbox[t]{2mm}{\multirow{4}{*}{\rotatebox[origin=c]{90}{\textbf{Cold Start}}}} & Llama-2 & 5.65 & 3.58 & 0.54 & 0.00 \\
& Mixtral & 27.96 & 12.77 & 4.62 & 3.28 \\
& Llama-3 & 59.14 & 42.98 & 10.26 & 7.75 \\
& GPT-4 & 58.60 & 41.65 & 34.25 & 37.94 \\
\hline
\end{tabular}
}
\caption{Gold leaderboard ablation evaluation (\%) on leaderboard dataset with paper threshold two.}
\label{tab:leaderboard_results_ablation}
\end{table}

\begin{table}
\centering
\renewcommand{\arraystretch}{1.1}
\resizebox{0.75\columnwidth}{!}{
\begin{tabular}{cccccc}
\cline{2-6}
& \textbf{Model} & \textbf{LR} &\textbf{PC} & \textbf{RC} & \textbf{AO} \\
\hline
\parbox[t]{2mm}{\multirow{4}{*}{\rotatebox[origin=c]{90}{\textbf{Partially}}}} & Llama-2 & 31.58 & 16.08 & 0.00 & 0.00 \\
& Mixtral & 31.58 & 17.22 & 9.59 & 14.62 \\
& Llama-3 & 73.68 & 48.07 & 12.11 & 10.25 \\
& GPT-4 & 63.16 & 39.60 & 31.29 & 38.67 \\
\hline
\parbox[t]{2mm}{\multirow{4}{*}{\rotatebox[origin=c]{90}{\textbf{Cold Start}}}} & Llama-2 & 8.77 & 5.26 & 1.75 & 0.00 \\
& Mixtral & 40.35 & 21.90 & 10.26 & 8.74 \\
& Llama-3 & 64.91 & 48.86 & 15.94 & 8.49 \\
& GPT-4 & 59.65 & 42.21 & 33.19 & 38.54 \\
\hline
\end{tabular}
}
\caption{Leaderboard evaluation only on masked leaderboards from partially masking setting (\%) on the leaderboard dataset with paper threshold two. Standard deviation for cold start are given in Table \ref{tab:cold_start_std_ablation_2}.}
\label{tab:masked_leaderbaord_results_ablation}
\end{table}

\section{Cold Start Results}\label{app:std_dev}

We provide standard deviation values for cold start experiments in Tables \ref{tab:cold_start_std_1}, \ref{tab:cold_start_std_2}, \ref{tab:cold_start_std_ablation_1} and \ref{tab:cold_start_std_ablation_2}.

\begin{table}
\centering
\setlength{\tabcolsep}{5pt}
\resizebox{\columnwidth}{!}{
\begin{tabular}{ccccc}
\hline
\textbf{Model} & \textbf{LR} & \textbf{PC} & \textbf{RC} & \textbf{AO} \\
\hline
Llama-2 & 16.05 $\pm$ 4.28 & 5.66 $\pm$ 1.76 & 0.49 $\pm$ 0.86 & 0.05 $\pm$ 0.09 \\
Mixtral & 49.38 $\pm$ 7.71 & 20.49 $\pm$ 5.74 & 8.10 $\pm$ 3.28 & 3.03 $\pm$ 3.47 \\
Llama-3 & 79.01 $\pm$ 14.02 & 58.78 $\pm$ 15.10 & 17.18 $\pm$ 4.72 & 17.63 $\pm$ 3.17 \\
GPT-4 & 81.48 $\pm$ 14.81 & 58.74 $\pm$ 7.42 & 46.13 $\pm$ 3.10 & 48.15 $\pm$ 4.42 \\
\hline
\end{tabular}
}
\caption{Mean and standard deviation values for cold start setting results in Table \ref{tab:leaderboard_results}.}
\label{tab:cold_start_std_1}
\end{table}

\begin{table}
\centering
\setlength{\tabcolsep}{5pt}
\resizebox{\columnwidth}{!}{
\begin{tabular}{ccccc}
\hline
\textbf{Model} & \textbf{LR} & \textbf{PC} & \textbf{RC} & \textbf{AO} \\
\hline
Llama-2 & 10.00 $\pm$ 0.00 & 3.10 $\pm$ 1.01 & 0.67 $\pm$ 1.15 & 0.00 $\pm$ 0.00 \\
Mixtral & 53.33 $\pm$ 15.27 & 22.67 $\pm$ 7.78 & 10.97 $\pm$ 3.22 & 7.91 $\pm$ 7.98 \\
Llama-3 & 86.67 $\pm$ 23.09 & 78.14 $\pm$ 16.61 & 22.59 $\pm$ 2.29 & 14.79 $\pm$ 3.34 \\
GPT-4 & 83.33 $\pm$ 11.55 & 60.76 $\pm$ 5.70 & 43.11 $\pm$ 4.00 & 37.50 $\pm$ 3.44 \\
\hline
\end{tabular}
}
\caption{Mean and standard deviation values for cold start setting results in Table \ref{tab:masked_leaderbaord_results}.}
\label{tab:cold_start_std_2}
\end{table}

\begin{table}
\centering
\setlength{\tabcolsep}{5pt}
\resizebox{\columnwidth}{!}{%
\begin{tabular}{ccccc}
\hline
\textbf{Model} & \textbf{LR} & \textbf{PC} & \textbf{RC} & \textbf{AO} \\
\hline
Llama-2 & 5.65 $\pm$ 1.14 & 3.58 $\pm$ 0.74 & 0.54 $\pm$ 0.47 & 0.00 $\pm$ 0.00 \\
Mixtral & 27.96 $\pm$ 4.93 & 12.77 $\pm$ 2.63 & 4.62 $\pm$ 1.72 & 3.28 $\pm$ 0.64 \\
Llama-3 & 59.14 $\pm$ 4.06 & 42.98 $\pm$ 7.47 & 10.26 $\pm$ 2.29 & 7.75 $\pm$ 2.98 \\
GPT-4 & 58.60 $\pm$ 1.86 & 41.65 $\pm$ 3.32 & 34.25 $\pm$ 1.49 & 37.94 $\pm$ 1.60 \\
\hline
\end{tabular}
}
\caption{Mean and standard deviation values for cold start setting results in Table \ref{tab:leaderboard_results_ablation}.}
\label{tab:cold_start_std_ablation_1}
\end{table}

\begin{table}
\centering
\setlength{\tabcolsep}{5pt}
\resizebox{\columnwidth}{!}{
\begin{tabular}{ccccc}
\hline
\textbf{Model} & \textbf{LR} & \textbf{PC} & \textbf{RC} & \textbf{AO} \\
\hline
Llama-2 & 8.77 $\pm$ 8.04 & 5.26 $\pm$ 4.56 & 1.75 $\pm$ 1.52 & 0.00 $\pm$ 0.00 \\
Mixtral & 40.35 $\pm$ 10.96 & 21.90 $\pm$ 5.28 & 10.26 $\pm$ 2.54 & 8.74 $\pm$ 0.58 \\
Llama-3 & 64.91 $\pm$ 8.04 & 48.86 $\pm$ 7.19 & 15.94 $\pm$ 1.46 & 8.49 $\pm$ 0.63 \\
GPT-4 & 59.65 $\pm$ 3.04 & 42.21 $\pm$ 7.43 & 33.19 $\pm$ 2.38 & 38.54 $\pm$ 3.20 \\
\hline
\end{tabular}
}
\caption{Mean and standard deviation values for cold start setting results in Table \ref{tab:masked_leaderbaord_results_ablation}.}
\label{tab:cold_start_std_ablation_2}
\end{table}

\section{Prompts} \label{app:prompts}

Table \ref{tab:tdmr_prompt} shows the system prompt for TDMR extraction phase. In Table \ref{tab:base_norm_prompt}, we provide prompt structure of \textit{fully pre-defined TDM triples} normalization setting. Tables \ref{tab:part_mask_first_prompt} and \ref{tab:part_mask_second_prompt} show the prompts of two steps of \textit{partially pre-defined TDM triples} normalization.

\begin{table*}
\centering
\small
\renewcommand{\arraystretch}{1.5}
\begin{tabular}{p{0.95\textwidth}}
\hline
You will be given several parts of a research paper as input. Please extract different tuples including the name of the task addressed in the paper, utilized datasets and evaluation metrics and corresponding results. Extract these tuples for only the best results obtained by proposed methods of the paper not baselines. Please use json format for each different tuple. Example format: [\{{"Task": "Task name", "Dataset": "Dataset name", "Metric": "Metric name", "Result": "Result score"}\}]. Your answer will immediately start with the json object satisfying the given template and contain nothing else. \\
\hline
\end{tabular}
\caption{System prompt for TDMR extraction from scientific papers.}
 \label{tab:tdmr_prompt}
\end{table*}

\begin{table*}
\centering
\small
\renewcommand{\arraystretch}{1.5}
\begin{tabular}{p{0.95\textwidth}}
\hline
You will be given a list of items. Then, an input will be provided. You will match the input with one of the items in the list. Your answer will ONLY consist of the matched item in the list, do not provide further explanations. If none of the items matches, say None. \\
Item list: \{'Combinatory Categorial Grammar (CCG) Supertagging', 'Constituency Parsing', 'Dependency Parsing', 'Dialogue Act Classification', 'Dialogue Generation', 'Entity Typing', 'Intent Detection and Slot Filling', 'Language Modeling', 'Linguistic Acceptability', 'Machine Translation', 'Named Entity Recognition (NER)', 'Natural Language Inference (NLI)', 'Paraphrase Detection', 'Part-of-Speech (POS) Tagging', 'Question Answering', 'Question Generation', 'Relation Classification', 'Response Generation', 'Sentiment Analysis', 'Summarization', 'Text Chunking', 'Text Similarity', 'Word Sense Induction'\} \\
Input: POS Tagging \\
\hline
\end{tabular}
\caption{Example prompt for \textit{fully pre-defined TDM triples} normalization setting of LLM outputs in terms of task names}
 \label{tab:base_norm_prompt}
\end{table*}

\begin{table*}
\centering
\small
\renewcommand{\arraystretch}{1.5}
\begin{tabular}{p{0.95\textwidth}}
\hline
You will be given a list of items. Then, an input entity will be provided. If the input entity matches one of the items in the list, your answer will be the matched item in the list. Else, output the entity without changing it. DO NOT make any other explanation. \\
Item list: \{'Combinatory Categorial Grammar (CCG) Supertagging', 'Constituency Parsing', 'Dependency Parsing', 'Dialogue Generation', 'Entity Typing', 'Language Modeling', 'Linguistic Acceptability', 'Machine Translation', 'Natural Language Inference (NLI)', 'Paraphrase Detection', 'Part-of-Speech (POS) Tagging', 'Question Answering', 'Question Generation', 'Relation Classification', 'Response Generation', 'Sentiment Analysis', 'Summarization', 'Text Chunking', 'Text Similarity', 'Word Sense Induction'\} \\
Input: Named entity recognition \\
\hline
\end{tabular}
\caption{Example prompt for first step of \textit{partially pre-defined TDM triples} normalization of LLM outputs in terms of task names. Different from the input set in Table \ref{tab:base_norm_prompt}, "Named Entity Recognition (NER)", "Intent Detection and Slot Filling" and "Dialogue Act Classification" tasks have been masked.} 
 \label{tab:part_mask_first_prompt}
\end{table*}

\begin{table*}
\centering
\small
\renewcommand{\arraystretch}{1.5}
\begin{tabular}{p{0.95\textwidth}}
\hline
You will be given a list of tuples. Then, an input tuple will be provided. If the input tuple matches one of the items in the list, your answer will be the matched item in the list. Else, output the tuple without changing it. Only output answer, DO NOT make any other explanation.\\
Item list: \{(Named Entity Recognition (NER), WNUT-16 - English, F1), (Intent Detection and Slot Filling, ATIS, Accuracy), (Summarization, CNN/DailyMail, ROGUE-1), (Word Sense Induction, SemEval 2013 Task 13, Fuzzy normalized mutual information (FNMI)), ...., (Machine Translation, WMT’14 EN-FR, BLEU), (Intent Detection and Slot Filling, SNIPS, F1), (Summarization, Gigaword, ROGUE-2), (Named Entity Recognition (NER), Ontonotes v5 - English, F1)\} \\
Input: (Entity Typing, WNUT-17, F1) \\
\hline
\end{tabular}
\caption{Example prompt for second step of \textit{partially pre-defined TDM triples} normalization of LLM outputs in terms of task names. After the simulation of the newly introduced tasks or dataset, this prompt check whether given input matches with any leaderboard tuples to implement proper evaluation. Due to convenient demonstrations purposes, not all tuples in gold leaderboards are listed.} 
 \label{tab:part_mask_second_prompt}
\end{table*}

\end{document}